\documentclass[conference]{IEEEtran}
\usepackage[T1]{fontenc}
\usepackage[utf8]{inputenc}
\usepackage{microtype}
\usepackage{graphicx}
\usepackage{amsmath,amssymb,amsthm}
\usepackage{booktabs,array,multirow,makecell}
\usepackage[numbers,sort&compress]{natbib}
\usepackage{xcolor}
\usepackage{needspace}
\usepackage{tikz}
\usetikzlibrary{arrows.meta,positioning}
\usepackage{pgfplots}
\pgfplotsset{compat=1.18}
\usepgfplotslibrary{groupplots}
\usepackage{xurl}
\usepackage[hidelinks]{hyperref}

\definecolor{auditblue}{HTML}{284D68}
\definecolor{auditteal}{HTML}{327D7D}
\definecolor{auditlight}{HTML}{EEF4F6}
\newcommand{\tablestyle}{\small\renewcommand{\arraystretch}{1.1}}
\newcommand{\compacttablestyle}{\footnotesize\renewcommand{\arraystretch}{1.1}}
\newcommand{\tablenote}[1]{\par\smallskip{\footnotesize\raggedright #1\par}}
\newcommand{\tablegroup}[1]{\textbf{#1}}
\theoremstyle{plain}
\newtheorem{theorem}{Theorem}

\theoremstyle{definition}

\newtheorem{remark}{Remark}
\newtheorem{finding}{Finding}
\AddToHook{env/finding/before}{\Needspace{4\baselineskip}}
\newcommand{\Lat}{\mathcal{L}}

\newcommand{\Pl}{\mathcal{P}}
\newcommand{\Act}{\mathcal{A}}
\newcommand{\Dcl}{\Delta^{\mathrm{cl}}}
\newcommand{\E}{\mathbb{E}}
\newcommand{\1}{\mathbf{1}}
\newcommand{\pp}{\,pp}

\title{When Does Better Confidence Improve a Knowledge-Graph Audit?}
\author{\IEEEauthorblockN{
Zehao Lin\IEEEauthorrefmark{1}, Kai Chen\IEEEauthorrefmark{1}, Xixuan Hao\IEEEauthorrefmark{1}\IEEEauthorrefmark{3}, Erenkai Zhu\IEEEauthorrefmark{2},\\
Chunyu Li\IEEEauthorrefmark{1}, Helan Hu\IEEEauthorrefmark{1}, Feiyu Xiong\IEEEauthorrefmark{1}}
\IEEEauthorblockA{\IEEEauthorrefmark{1}MemTensor (Shanghai) Technology Co., Ltd.\\
\IEEEauthorrefmark{2}Shanghai Jiaotong University\\
\IEEEauthorrefmark{3}The Hong Kong University of Science and Technology (Guangzhou)\\
\texttt{\{linzh,chenk,licy,huhl,xiongfy\}@memtensor.cn}\\
\texttt{xhao390@connect.hkust-gz.edu.cn}\quad\texttt{erz21@sjtu.edu.cn}}}
\hypersetup{pdftitle={When Does Better Confidence Improve a Knowledge-Graph Audit?},
pdfauthor={Zehao Lin; Kai Chen; Xixuan Hao; Erenkai Zhu; Chunyu Li; Helan Hu; Feiyu Xiong}}

\begin{document}
\bstctlcite{IEEEReferenceSettings}
\maketitle
\pagestyle{plain}
\thispagestyle{plain}
\begin{abstract}
Confidence estimates guide human review in LLM-built knowledge graphs, but an erroneous shared entity need not be the only reason an incident fact is wrong. We study how this distinction affects budgeted audits, alongside selection headroom and label validity. Modelling non-adaptive auditing as budgeted covering, we characterise the minimal decision-relevant error statistics and bound the regret of scalar-preserving first-order planning by a structural mass $\Dcl$. We simulate ideal repair against benchmark-aligned labels for five extraction configurations and two cross-document graphs on Re-DocRED, followed by frozen confirmation on test and replication on 500 hash-selected training documents. An error-rate $\times$ fan-out baseline leaves 3.98--4.74 percentage points more errors than joint planning with the same actions. Independent-block planning also avoids this disadvantage: its held-out point estimates improve on a fact-only scalar baseline by 1.22--1.79 points and are slightly better than the joint model, although the frozen equivalence criterion remains inconclusive. On 300 DWIE training documents, independent and joint plans also outperform fan-out in point estimates, with extractor-dependent gains over fact-only review. For verification, a generic support check improves three of four model--budget configurations on each held-out set. These comparisons motivate evaluating complete-repair probabilities and retaining both fact-only and factorised baselines when choosing an audit model.
\end{abstract}

\section{Introduction}
\label{sec:intro}

\begin{figure*}[t]
\centering
\includegraphics[width=\textwidth]{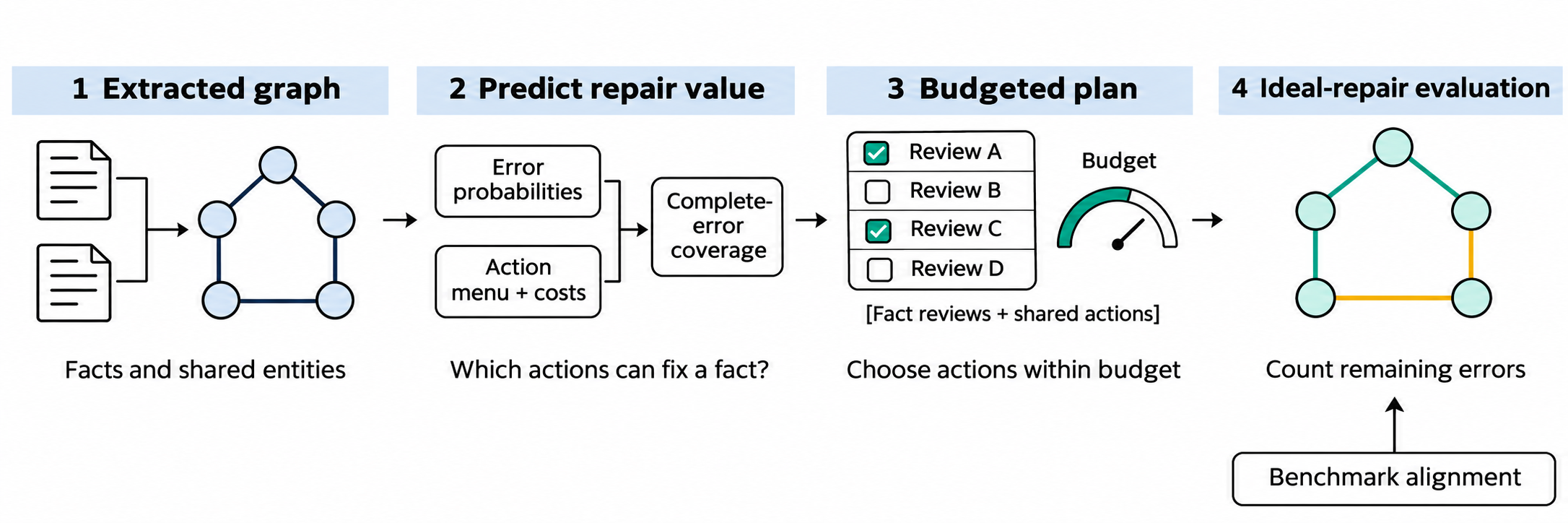}
\caption{The audit evaluation pipeline. Error predictions and a fixed action menu determine a plan under the review budget. Evaluation simulates ideal repair against benchmark-aligned labels: fact $i$ is correct after review exactly when the chosen actions cover its complete error set, $E_i\subseteq\tau_i(\Pl)$.}
\label{fig:overview}
\end{figure*}

Knowledge-graph curation must allocate limited review effort across facts and shared entities. Accuracy estimation measures graph quality \citep{gao-etal-2019-efficient,marchesin-silvello-2024-efficient}; question selection and data repair use dependencies to decide where intervention helps \citep{whang-etal-2013-question,zhang-etal-2013-reducing,rekatsinas-etal-2017-holoclean}. When graphs are extracted by LLMs, confidence scores provide an additional signal for these decisions. We ask when a better error predictor actually reduces the errors remaining after a budgeted audit.

LLMs generate relation records \citep{wadhwa-etal-2023-revisiting} and assemble knowledge graphs \citep{mo-etal-2025-kggen}, but their outputs require substantial review. The five configurations in our Re-DocRED study reach precision at most 0.53 on verifiable predictions and end-to-end recall below 0.23 (Table~\ref{tab:populations}), consistent with prompted-extraction results on this benchmark family \citep{peng-etal-2023-when,li-etal-2024-lmrc,xue-etal-2024-autore}. Unverifiable predictions account for 10.2--18.8\% of outputs; counting them all as wrong lowers maximum precision to 0.461. Document extraction systems use reliability scores to allocate human review \citep{roy-etal-2026-confbench,goh-mueller-2026-construct,kumar-2026-extractconf}.

Shared entities suggest modelling errors jointly, since one entity review may repair several facts; related benefits arise in crowd entity resolution and schema matching \citep{whang-etal-2013-question,zhang-etal-2013-reducing}. The distinction between error risk and intervention value has also been studied for generated answers \citep{park-etal-2026-risk}. In our KG setting, a marginal-risk baseline credits a shared review according to error probability times fan-out. A factorised alternative multiplies the probabilities of the component states, assigning repair value only when the remaining components are correct. Comparing both with joint planning separates the choice of repair objective from the need to model statistical dependence.

Beyond representations, we examine verification gains across review budgets and the benchmark labels used to measure them. \emph{Repairability} asks which errors an available action can fix: correcting a shared entity may leave a fact wrong if its relation is also wrong. \emph{Selection headroom} asks how much a better ranking can change the audit: a fact-review queue consisting entirely of errors already uses every unit of budget productively. \emph{Label validity} asks whether the benchmark distinguishes extraction errors from annotation omissions.

Our analysis maps an action menu to the minimal error statistics needed for planning. It identifies when a per-fact scalar suffices, bounds scalar-preserving first-order regret by the measurable structural mass $\Dcl$, and relates ranking improvement to the errors a budgeted queue can still recover.

The empirical study compares six uncertainty representations under shared features, folds, model families, costs and planning procedures. This design separates a representation change at fixed actions from adding shared actions to a fixed planner. Development experiments cover five extraction configurations and two cross-document graphs. We then freeze three hypotheses before held-out alignment and evaluate them on Re-DocRED test and a hash-selected training sample. The fan-out disadvantage is confirmed on both sets. Further analyses examine DWIE, repair failures, verification and label sensitivity.

\section{The audit as budgeted covering}
\label{sec:setting}

An extractor makes, per document, a set of \emph{decisions}. These are entity decisions (which mentions form an entity and, across documents, which KB entity it links to) and one local decision per fact (relation and direction). Fact $i$ depends on the decisions $S_i$, has weight $w_i$ ($=1$ here), and has an \emph{error set} $E_i\subseteq S_i$ of wrong decisions, with law $P_i(T)=P(E_i=T\mid X)$. An \emph{action} $a$ costs $c_a$. It either has a global coverage $D(a)$ (an entity review repairs that decision wherever it is used) or a local scope $\sigma(a)=i$ (a whole-fact review). A plan $\Pl$ covers $\tau_i(\Pl)=S_i$ if it reviews fact $i$ as a whole, and $S_i\cap\bigcup_{a\in\Pl}D(a)$ otherwise.

We assume (A1) ideal repair, i.e.\ fact $i$ is correct after the audit iff $E_i\subseteq\tau_i(\Pl)$; (A2) non-adaptive plans, chosen before any review outcome is seen (reviews do not adapt to observed outcomes); (A3) an additive objective; and (A4) an action menu that does not depend on $P$. The planner then maximises, under a budget,
\begin{equation}
F_P(\Pl)=\sum_i w_i\,\Phi_i(\tau_i(\Pl)),
\label{eq:objective}
\end{equation}
where $\Phi_i(U)=\sum_{T\subseteq U}P_i(T)=P(E_i\subseteq U)$ is the cumulative error-set probability.\footnote{For $p_i=P(E_i=\varnothing)<1$, the conditional quantity $(\Phi_i(U)-p_i)/(1-p_i)$ is a normalized belief function in the sense of \citet{shafer-1976-mathematical}. The unconditional $\Phi_i$ also includes initially correct facts.} A deleted wrong fact counts as repaired, so KG precision depends on the repair model (\S\ref{sec:verif}).

\begin{figure}[t]
\centering
\begin{tikzpicture}[
  x=1cm,y=1cm,
  state/.style={draw=auditblue!65,fill=auditlight,rounded corners=2pt,
    minimum width=0.95cm,minimum height=0.48cm,inner sep=4pt,font=\small},
  edge/.style={-{Stealth[length=4pt]},draw=auditblue,line width=0.7pt}]
  \node[state] (e) at (2,0) {$\varnothing$};
  \node[state] (h) at (0.7,1) {$\{h\}$};
  \node[state] (t) at (3.3,1) {$\{t\}$};
  \node[state] (ht) at (2,2) {$\{h,t\}$};
  \node[state,fill=auditteal!10,draw=auditteal] (s) at (2,3) {$S_i=\{h,t,\ell\}$};
  \draw[edge] (e) -- (h);
  \draw[edge] (e) -- (t);
  \draw[edge] (h) -- (ht);
  \draw[edge] (t) -- (ht);
  \draw[edge] (ht) -- (s);
  \draw[edge,dashed,draw=auditteal] (e.east) -- (4.65,0) --
    node[right,align=left,font=\footnotesize,text=auditteal]{whole-fact\\review} (4.65,3) -- (s.east);
\end{tikzpicture}
\caption{Reachable coverage for one fact: $h$ and $t$ are entity decisions and $\ell$ is the local relation decision. Solid arrows show the cover relations; whole-fact review reaches $S_i$ directly (dashed). The decision-relevant \emph{closure variable} $Z_i$ takes values in the five reachable covered sets $\{\varnothing,h,t,ht,S\}$.}
\label{fig:lattice}
\end{figure}
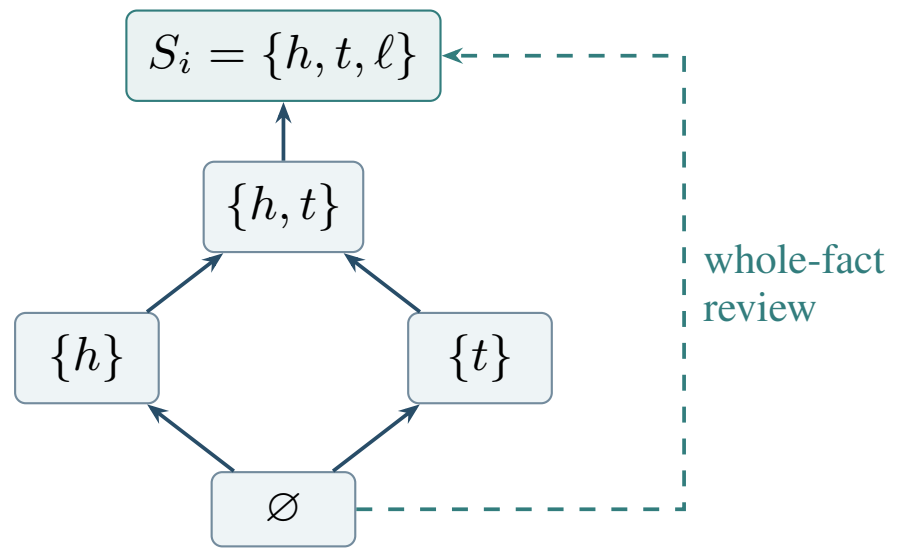

\section{What the audit decision needs}
\label{sec:theory}

The audit objective determines which features of the error distribution are needed for planning. Let $\Lat_i=\{\tau_i(\Pl):\Pl\subseteq\Act\}$ be the \emph{action lattice} of reachable covered sets. It contains $\varnothing$ and is closed under union. Two decisions are in the same \emph{block} if every action covers both or neither.

\begin{theorem}[Decision sufficiency]
\label{thm:suff}
Under (A1)--(A4):
(i) the optimal plan, for every cost vector and budget, depends on $P$ only through $g_i(U)=P(\varnothing\neq E_i\subseteq U)$, $U\in\Lat_i\setminus\{\varnothing\}$;
(ii) no coordinate can be dropped;
(iii) only with an outside option of $P$-independent value (admit or discard) does $p_i$ also become decision-relevant, and then $\Phi_i|_{\Lat_i}$ is minimal;
(iv) the scalar $p_i=P(E_i=\varnothing)$ suffices for all instances \emph{iff} $\Lat_i\subseteq\{\varnothing,S_i\}$, i.e.\ every action touching fact $i$ covers all of it;
(v) the minimal dimension is $|\Lat_i|-1$.
\end{theorem}

\begin{proof}[Proof sketch]
\emph{(i)} Write $\Phi_i(U)=p_i+g_i(U)$ for $U\neq\varnothing$. Then $F_P(\Pl)=\sum_iw_ip_i+\sum_iw_ig_i(\tau_i(\Pl))$. The first term does not depend on the plan, so the ranking of plans depends on $P$ only through $g$.

\emph{(ii)} Let $\Pl_0$ reach $U_0$ for fact $i$. Give each action of $\Pl_0$ the cost $1/|\Pl_0|$, give every other action the cost 2, and set the budget to 1. Add a competing fact $j$ with a private decision, a single action of cost 1 and gain $w_jq$. Then $\Pl_0$ is optimal iff $w_ig_i(U_0)>w_jq$. Choosing $q$ between the two models' values of $w_ig_i(U_0)/w_j$ separates them. The coordinates are affinely independent, because the zeta matrix restricted to $\Lat_i$ is unit upper-triangular.

\emph{(iii)} An outside option with a $P$-independent value compares against $F$ itself, so $p_i$ becomes decision-relevant.

\emph{(iv)} If some action has $\varnothing\neq D(a)\cap S_i\neq S_i$, then by (ii) the coordinate $g_i(D(a)\cap S_i)$ is decision-relevant, and it is not a function of $p_i$. Conversely, if $\Lat_i\subseteq\{\varnothing,S_i\}$, then $g_i(S_i)=1-p_i$.

\emph{(v)} $\Lat_i$ is the union closure of the generators $D(a)\cap S_i$ (and $S_i$ when fact review exists). It need not contain every union of blocks, so the dimension is $|\Lat_i|-1$, not $2^{m_i}-1$.
\end{proof}

Part (i) is linearity of \eqref{eq:objective} in $(\1[E\subseteq U])_{U\in\Lat_i}$, connecting the audit objective to property elicitation \citep{frongillo-kash-2021-elicitation} and calibration of queried events \citep{kuleshov-liang-2015-calibrated}. For intersection-closed lattices, the closure $Z_i=\bigcap\{U\in\Lat_i:E_i\subseteq U\}$ carries the same information; in Figure~\ref{fig:lattice} it has 5 classes.

\paragraph{Value of joint modelling}
Let $B(E_i)$ be the set of blocks that $E_i$ touches, $m_i(b)=P(B(E_i)=\{b\})$ the first-order masses, and $\Delta=\sum_iw_iP(|B(E_i)|\ge2)$ the \emph{raw multi-block mass}. A closure class is \emph{union-reducible} if it is the union of strictly smaller lattice elements. The \emph{pure-structure joint mass} is $\Dcl=\sum_iw_iP(Z_i\text{ union-reducible})\le\Delta$.

\begin{theorem}[Worst-case value of joint modelling]
\label{thm:delta}
(i) With first-order gain $G_1$, $G_1\le G\le G_1+\Delta$ for every plan, so first-order planning has regret $\le\Delta$.
(ii) Bound (i) is tight up to a factor of 2 for deterministic planners that see only first-order statistics.
(iii) If every $\Lat_i$ is intersection-closed and contains $S_i$, the \emph{scalar-preserving first-order} planner, which moves union-reducible mass to $S_i$, has regret $\le\Dcl$. This bound is tight up to a factor of 2.
\end{theorem}

\begin{proof}[Proof sketch]
\emph{(i)} $G-G_1=\sum_iw_i\sum_{T:|B(T)|\ge2}P_i(T)\1[T\subseteq\tau_i]\in[0,\Delta]$. With $\Pl_1=\arg\max G_1$: $G(\Pl^*)\le G_1(\Pl^*)+\Delta\le G_1(\Pl_1)+\Delta\le G(\Pl_1)+\Delta$.

\emph{(ii)} Place two pairs of disjoint two-block error sets, each with mass $\delta/2$, and interleave them across two instances so that all first-order statistics coincide. Any fixed plan collects the multi-block gain in at most one of the two instances.

\emph{(iii)} For $U\in\Lat_i$, $E_i\subseteq U\iff Z_i\subseteq U$. So $G-G_s=\sum_iw_i\sum_{Z\text{ union-red.}}P(Z_i=Z)\1[Z\subseteq\tau_i,\tau_i\neq S_i]\in[0,\Dcl]$, and the three-step argument of (i) applies. To see $\Dcl\le\Delta$: the closure of a single block cannot be union-reducible, because every lattice element is a union of blocks.
\end{proof}

The bounds compare static planners using the true error law. A small $\Dcl$ limits the possible structural advantage over the scalar-preserving planner; a large value leaves room for such an advantage. In Figure~\ref{fig:lattice}, the only union-reducible class is $\{h,t\}$, whereas an error involving both a link and the local relation closes to the whole fact. Such errors contribute to raw $\Delta$ alone.

\begin{remark}[Why node scalars can hurt]
\label{rem:marginal}
Ranking entity reviews by ``node error $\times$ fan-out'' plans with the marginals $\mu_i(b)=P(b\in E_i)$ in place of $m_i(b)$. The difference $\mu_i(b)-m_i(b)=P(b\in E_i,|B(E_i)|\ge2)$ is exactly the mass that a node review \emph{cannot} repair alone. This identity identifies the repair credit assigned to facts that require an additional action; the aggregate plug-in bound is $(1+2m_{\max})\Delta$, where $m_{\max}$ is the largest number of blocks per fact.
\end{remark}

\paragraph{Scoring complete repairs}
Consider two entity-review candidates with equal cost and fan-out one. The first fact has error set $\{h,\ell\}$ with probability 0.6 and is otherwise correct; the second has error set $\{h\}$ with probability 0.4 and is otherwise correct. Node risk ranks the first review higher, although repairing its head alone removes zero errors in expectation, compared with 0.4 for the second. Under Figure~\ref{fig:lattice}'s action menu, both facts have $\Dcl=0$: the first error closes to the whole fact. The example shows that correcting the node queue's accounting can be valuable even when the structural regret bound vanishes.

\paragraph{Selection headroom for fact-level scores}
With fact-only unit-cost plans and $k=\lfloor bN\rfloor$ reviews, the clairvoyant plan removes $\min(k,\#\text{wrong})$ errors and the top $k$ facts of a score remove $k\cdot\mathrm{prec@}b$, where $\mathrm{prec@}b$ is their share of wrong facts. Any better fact-level score can thus gain at most $\min(k,\#\text{wrong})-k\cdot\mathrm{prec@}b$ errors, which is $k(1-\mathrm{prec@}b)$ while $k\le\#\text{wrong}$; we call this the \emph{selection gap}. When $k\le\#\text{wrong}$, the gap equals the number of correct facts reviewed. With entity actions, comparison to the fact-only oracle serves as a reference gap; the bound above applies to fact-only plans. The value of one verification call has the one-step exchange form of the expected value of sample information (Section~\ref{sec:practical}; \citep{strong-etal-2015-estimating}): it is non-zero only if the call can move a fact across the cutoff.

\paragraph{Diagnostics}
\label{sec:diagnostics}
Labelled development data support three complementary diagnostics: $\Dcl$ quantifies the structural regret bound; precision@budget determines the available improvement in fact-only ranking; and the oracle action gain measures how much additional actions help under perfect information. The last quantity evaluates the action menu, while the first two concern the information used to select among its actions.

\paragraph{Scope of the structural bound}
The structural bound applies to action menus that satisfy ideal repair and intersection closure. Across documents, fact review, entity review and shared-node relinking satisfy these conditions (Section~\ref{sec:setup}). Shared-node relabelling can break a correct member and, when a fact's endpoints link to the same node, destroy intersection closure. Evaluating menus with such actions therefore requires explicit accounting for both repairs and damage.

\section{Experimental design}
\label{sec:setup}

\paragraph{Data and extraction}
Development experiments use the 500 documents and 17,284 gold facts of Re-DocRED dev \citep{tan-etal-2022-revisiting}, a relabelling of DocRED \citep{yao-etal-2019-docred} with 96 relations. Four models extract entities, coreference and facts end to end: Claude Sonnet~5, GPT-4.1, GPT-4.1-mini and Qwen3.5-27B. The fifth configuration, \emph{Qwen-R}, uses the same Qwen model with a reasoning-first prompt. We collect verbal confidences and available token log-probabilities (Section~\ref{sec:implementation}). Unparsed documents count as misses: Qwen-R fails on 41 documents, so its higher precision than direct Qwen (0.510 versus 0.421) accompanies lower end-to-end recall (0.200 versus 0.208).

\paragraph{Labels and audit population}
Aligning predicted and gold mentions yields an error mask over $\{h,t,\ell\}$ for each fact, where entity errors correspond to merges of distinct gold entities. Our primary \emph{lenient} policy treats selected apparent gold splits, such as acronym variants, as benign, whereas the \emph{strict} policy counts all merges as a sensitivity check. A fact is \emph{verifiable} when both endpoints align to gold entities. The remaining 10.2--18.8\% of predictions are excluded from the audit population and budget denominator, with their effect on precision reported in Table~\ref{tab:populations}. Repair outcomes are evaluated against this alignment and label policy.

\paragraph{Cross-document graphs}
Qwen and GPT-4.1 extractions form the two cross-document graphs (498 documents). GPT-4.1-mini links predicted entities to Wikipedia, and Linked-DocRED provides gold links \citep{genest-etal-2023-linked}. Available actions are fact review, document-entity review and relink, each costing one unit; relabel costs $1+0.1$ per member and split costs $1+0.5$ per member.

\paragraph{Models and planning}
We compare fact scalars, node marginals, independent blocks, direct utility regression, the lattice/joint representation, and an 8-class error-mask representation. The lattice has 5 classes within documents and 37 across documents. Each setting uses the same available features across representations: 86 extraction features within documents, augmented by linking and canonicalisation features across documents. Logistic regression (LR) and gradient boosting (GB) share a five-fold document-grouped development split. Label-free rankers provide an additional comparison in Section~\ref{sec:verif}.

Across documents, the independent model fits five field predictors: binary head-entity, tail-entity and local-error models, plus two three-state link models (correct, label error, membership error). Their product gives the repair-event probabilities; coincident endpoints are counted once. The joint model instead predicts 37 closure classes. These models use the same fact features, but neither is guaranteed to be better calibrated.

The planner maximises the same repair objective under each representation, using the solvers described in Section~\ref{sec:implementation}. Budget $b$ is a fraction of the cost of reviewing every verifiable fact. We report residual benchmark errors as counts or percentage points (pp) of initially wrong facts.

\paragraph{Statistical analysis}
Table~\ref{tab:confirm} reports the pre-specified held-out hypothesis tests; development comparisons, sensitivity analyses and secondary benchmarks are diagnostic. Development headline intervals use a re-planned document-cluster bootstrap with 100 resamples: each plan is re-solved at the resample's budget, with fold models held fixed. Other development comparisons use a budget-normalised, fixed-plan approximation with 2,000 resamples. Development equivalence tests use TOST \citep{schuirmann-1987-comparison} with margins of $\pm1\%$ of wrong facts within documents and $\pm0.5$\pp{} across documents, chosen after inspecting the development intervals. We additionally report the smallest passing margin $\delta^*$. The held-out study uses pre-specified halved margins, 1,000 re-planned resamples and Holm correction \citep{holm-1979-simple} (Section~\ref{sec:confirm}).

\begin{table*}[t]
\caption{Extraction populations and alignment-sensitive precision.}
\label{tab:populations}
\centering
\tablestyle
\setlength{\tabcolsep}{3.5pt}
\begin{tabular*}{\textwidth}{@{\extracolsep{\fill}}lrrrrrrrr@{}}
\toprule
Extraction config & ok/fail & verifiable & wrong & P & R & unverifiable (share) & P, unverif.\ wrong & P, unverif.\ right \\
\midrule
\multicolumn{9}{@{}l}{\textit{Single-document (Re-DocRED dev, 500 docs, lenient)}}\\
Claude Sonnet 5 (direct) & 494/6 & 7,349 & 3,468 & 0.528 & 0.223 & 1,064 (12.6\%) & 0.461 & 0.588 \\
GPT-4.1 (direct) & 500/0 & 8,440 & 5,850 & 0.307 & 0.148 & 1,550 (15.5\%) & 0.259 & 0.414 \\
GPT-4.1-mini (direct) & 499/1 & 5,472 & 3,974 & 0.274 & 0.086 & 624 (10.2\%) & 0.246 & 0.348 \\
Qwen3.5-27B (direct) & 499/1 & 8,696 & 5,039 & 0.421 & 0.208 & 1,463 (14.4\%) & 0.360 & 0.504 \\
Qwen3.5-27B (reason) & 459/41 & 6,818 & 3,339 & 0.510 & 0.200 & 1,583 (18.8\%) & 0.414 & 0.603 \\
\midrule
\multicolumn{9}{@{}l}{\textit{Cross-document (498 docs; links by GPT-4.1-mini)}}\\
Qwen3.5-27B & -- & 8,685 & 5,820 & \multicolumn{5}{l}{link accuracy 83.4\%} \\
GPT-4.1 & -- & 8,426 & 6,404 & \multicolumn{5}{l}{link accuracy 83.7\%} \\
\bottomrule
\end{tabular*}
\tablenote{P is precision over verifiable facts; R is end-to-end recall over all 17,284 dev gold facts, counting failed documents as misses. Wrong counts use lenient labels. Unverifiable predictions are excluded from audit budgets; the last two columns count all of them as wrong or right.}

\end{table*}

\subsection{Implementation and measurement}
\label{sec:implementation}
The extraction prompt supplies the 96 relation definitions and asks for entity clusters, mentions, facts, evidence sentences and confidence values. The direct variant returns JSON, whereas Qwen-R emits an analysis before JSON; extraction and linking both use temperature zero. The 86 baseline features comprise 36 token-log-probability, 28 entity, 13 structural, seven verbal-confidence and two document features, with relation frequencies computed over predictions. A verifier sees the numbered source, schema and candidate fact, without gold labels, extractor confidence, evidence IDs or reasoning. The generic verification check, denoted G, returns a yes/no support decision and a probability; failed checks leave verification features missing while retaining the fact.

Scikit-learn LR and GB share document-grouped GroupKFold splits, with GB routing missing features natively. HiGHS through SciPy solves within-document plans exactly, while cross-document plans use LP-and-fix when needed. Recorded oracle relative gaps are at most $1.2\times10^{-4}$ for Qwen and $7.4\times10^{-4}$ for GPT-4.1, and the full-menu solver accounts for members broken by relabel operations. We report the token cost of verification separately from the human-review budget.

Development extraction used an OpenAI-compatible gateway (24--26 September 2026) with \texttt{claude-sonnet-5}, \texttt{gpt-4.1}, \texttt{gpt-4.1-mini} and \texttt{qwen3.5-27b}. The gateway exposed no dated snapshots. Code, numerical features and reference outputs for reproducing the comparisons are available online.\footnote{\url{https://drive.google.com/drive/folders/1Jwx5sL_njTiCvyL3p1g8BzaSSGlifLki}}

\section{Representation and repair actions}
\label{sec:repr}

\begin{table*}[t]
\caption{Within-document development comparisons (lenient, GB, $b=0.2$).}
\label{tab:matrix}
\centering\tablestyle
\setlength{\tabcolsep}{2pt}
\begin{tabular*}{\textwidth}{@{\extracolsep{\fill}}lccccc@{}}
\toprule
Contrast & Claude & GPT-4.1 & mini & Qwen & Qwen-R \\
\midrule
Initially wrong facts & 3,468 & 5,850 & 3,974 & 5,039 & 3,339 \\[2pt]
Raw $\Delta$ (facts) & 213 (6.14\%) & 204 (3.49\%) & 165 (4.15\%) & 181 (3.59\%) & 151 (4.52\%) \\[2pt]
$\Dcl$ (facts) & 2 & 2 & 0 & 4 & 2 \\[2pt]
\midrule
(a) Oracle: + shared actions & \makecell{$+3.52$$^{*}$ {\footnotesize(+122)}\\{\footnotesize$[+2.37, +4.79]$}} & \makecell{$+0.96$$^{*}$ {\footnotesize(+56)}\\{\footnotesize$[+0.59, +1.71]$}} & \makecell{$+1.01$$^{*}$ {\footnotesize(+40)}\\{\footnotesize$[+0.46, +1.84]$}} & \makecell{$+1.65$$^{*}$ {\footnotesize(+83)}\\{\footnotesize$[+0.96, +2.40]$}} & \makecell{$+2.40$$^{*}$ {\footnotesize(+80)}\\{\footnotesize$[+1.23, +3.58]$}} \\[2pt]
(b) Learned: + shared actions & \makecell{$+0.03$ {\footnotesize(+1)}\\{\footnotesize$[-0.42, +0.44]$}} & \makecell{$+0.02$ {\footnotesize(+1)}\\{\footnotesize$[-0.05, +0.10]$}} & \makecell{$+0.00$ {\footnotesize(+0)}\\{\footnotesize$[+0.00, +0.00]$}} & \makecell{$+0.18$ {\footnotesize(+9)}\\{\footnotesize$[+0.00, +0.41]$}} & \makecell{$-0.03$ {\footnotesize(-1)}\\{\footnotesize$[-0.11, +0.00]$}} \\[2pt]
(c) Node: + shared actions & \makecell{$-0.89$$^{*}$ {\footnotesize(-31)}\\{\footnotesize$[-1.75, -0.26]$}} & \makecell{$-0.14$ {\footnotesize(-8)}\\{\footnotesize$[-0.35, +0.07]$}} & \makecell{$-0.40$$^{*}$ {\footnotesize(-16)}\\{\footnotesize$[-0.65, -0.10]$}} & \makecell{$+0.08$ {\footnotesize(+4)}\\{\footnotesize$[-0.31, +0.38]$}} & \makecell{$-0.21$ {\footnotesize(-7)}\\{\footnotesize$[-0.73, +0.18]$}} \\[2pt]
\makecell[l]{(s) Lattice + shared actions\\vs.\ fact scalar, fact only} & \makecell{$+0.52$ {\footnotesize(+18)}\\{\footnotesize$[-0.63, +1.10]$}} & \makecell{$+0.05$ {\footnotesize(+3)}\\{\footnotesize$[-0.08, +0.20]$}} & \makecell{$-0.15$ {\footnotesize(-6)}\\{\footnotesize$[-0.33, +0.00]$}} & \makecell{$+0.32$$^{*}$ {\footnotesize(+16)}\\{\footnotesize$[+0.04, +0.53]$}} & \makecell{$+0.18$ {\footnotesize(+6)}\\{\footnotesize$[-0.39, +0.73]$}} \\[2pt]
\midrule
(d) Lattice vs. node marginal & \makecell{$+1.41$$^{*}$ {\footnotesize(+49)}\\{\footnotesize$[+0.63, +1.92]$}} & \makecell{$+0.19$ {\footnotesize(+11)}\\{\footnotesize$[-0.07, +0.46]$}} & \makecell{$+0.25$ {\footnotesize(+10)}\\{\footnotesize$[-0.15, +0.62]$}} & \makecell{$+0.24$ {\footnotesize(+12)}\\{\footnotesize$[-0.13, +0.63]$}} & \makecell{$+0.39$ {\footnotesize(+13)}\\{\footnotesize$[-0.45, +1.12]$}} \\[2pt]
(e) Lattice vs. independent & \makecell{$+0.52$ {\footnotesize(+18)}\\{\footnotesize$[-0.29, +1.00]$}} & \makecell{$+0.10$ {\footnotesize(+6)}\\{\footnotesize$[-0.07, +0.27]$}} & \makecell{$-0.13$ {\footnotesize(-5)}\\{\footnotesize$[-0.39, +0.10]$}} & \makecell{$+0.06$ {\footnotesize(+3)}\\{\footnotesize$[-0.29, +0.38]$}} & \makecell{$+0.09$ {\footnotesize(+3)}\\{\footnotesize$[-0.49, +0.58]$}} \\[2pt]
(f) Lattice vs. direct utility & \makecell{$+0.61$ {\footnotesize(+21)}\\{\footnotesize$[-0.40, +1.10]$}} & \makecell{$+0.05$ {\footnotesize(+3)}\\{\footnotesize$[-0.08, +0.20]$}} & \makecell{$-0.15$ {\footnotesize(-6)}\\{\footnotesize$[-0.33, +0.00]$}} & \makecell{$+0.12$ {\footnotesize(+6)}\\{\footnotesize$[-0.17, +0.43]$}} & \makecell{$+0.21$ {\footnotesize(+7)}\\{\footnotesize$[-0.35, +0.73]$}} \\[2pt]
TOST $\delta^*$: (e) / (f) & 0.92 / 1.01$^{?}$ & 0.26 / 0.17 & 0.38 / 0.30 & 0.34 / 0.38 & 0.54 / 0.63 \\[2pt]
\midrule
(g) Capacity: LR to GB & \makecell{$+3.78$$^{*}$ {\footnotesize(+131)}\\{\footnotesize$[+2.51, +4.56]$}} & \makecell{$+0.68$$^{*}$ {\footnotesize(+40)}\\{\footnotesize$[+0.36, +0.98]$}} & \makecell{$+0.20$ {\footnotesize(+8)}\\{\footnotesize$[-0.03, +0.53]$}} & \makecell{$+0.85$$^{*}$ {\footnotesize(+43)}\\{\footnotesize$[+0.29, +1.29]$}} & \makecell{$+1.14$$^{*}$ {\footnotesize(+38)}\\{\footnotesize$[+0.06, +2.01]$}} \\[2pt]
(h) Gap to oracle & \makecell{$+7.38$$^{*}$ {\footnotesize(+256)}\\{\footnotesize$[+6.10, +8.93]$}} & \makecell{$+1.47$$^{*}$ {\footnotesize(+86)}\\{\footnotesize$[+0.94, +2.20]$}} & \makecell{$+1.48$$^{*}$ {\footnotesize(+59)}\\{\footnotesize$[+0.88, +2.32]$}} & \makecell{$+2.48$$^{*}$ {\footnotesize(+125)}\\{\footnotesize$[+1.71, +3.14]$}} & \makecell{$+6.95$$^{*}$ {\footnotesize(+232)}\\{\footnotesize$[+5.54, +8.72]$}} \\[2pt]
\bottomrule
\end{tabular*}
\tablenote{Changes are percentage points of initially wrong facts, with counts in parentheses; positive means fewer residual errors. Rows a--c add actions; d--f hold actions fixed. Brackets: 95\% re-planned bootstrap intervals ($R=100$, models fixed); $^{*}$ excludes zero. TOST margin: $\pm1\%$ of wrong facts; $^{?}$ inconclusive.}

\end{table*}

\begin{table*}[t]
\caption{Cross-document development comparisons (lenient, GB, $b=0.2$).}
\label{tab:matrix-cross}
\centering\tablestyle
\setlength{\tabcolsep}{2pt}
\begin{tabular*}{\textwidth}{@{\extracolsep{\fill}}lcc@{}}
\toprule
Contrast & Qwen & GPT-4.1 \\
\midrule
Initially wrong facts & 5,820 & 6,404 \\[2pt]
Raw $\Delta$ (facts) & 1642 (28.2\%)$^{\S}$ & 1863 (29.1\%)$^{\S}$ \\[2pt]
$\Dcl$ (facts) & 66 (1.1\%)$^{\S}$ & 54 (0.8\%)$^{\S}$ \\[2pt]
\midrule
(a) Oracle: + shared actions & $+6.89$$^{*}$\;$[+5.64, +8.16]$ & $+4.06$$^{*}$\;$[+3.11, +4.94]$ \\[2pt]
(b) Learned: + shared actions & $+1.46$$^{*}$\;$[+0.87, +1.98]$ & $+0.69$$^{*}$\;$[+0.28, +1.02]$ \\[2pt]
(c) Node: + shared actions & $-3.13$$^{*}$$^{\mathrm{N}}$\;$[-4.35, -1.92]$ & $-3.98$$^{*}$$^{\mathrm{N}}$\;$[-4.87, -3.02]$ \\[2pt]
\makecell[l]{(s) Lattice + shared actions\\vs.\ fact scalar, fact only} & $+1.49$$^{*}$\;$[+0.90, +2.07]$ & $+0.72$$^{*}$\;$[+0.33, +1.09]$ \\[2pt]
\midrule
(d) Lattice vs. node marginal & $+4.62$$^{*}$\;$[+3.77, +5.54]$ & $+4.70$$^{*}$\;$[+3.78, +5.43]$ \\[2pt]
(e) Lattice vs. independent & $+0.05$\;$[-0.28, +0.51]$ & $-0.09$\;$[-0.29, +0.12]$ \\[2pt]
(f) Lattice vs. direct utility & $+0.09$\;$[-0.13, +0.32]$ & $+0.14$\;$[-0.02, +0.26]$ \\[2pt]
TOST $\delta^*$: (e) / (f) & 0.47 / 0.27 & 0.26 / 0.24 \\[2pt]
\midrule
(g) Capacity: LR to GB & $+0.43$$^{\dagger}$\;$[-0.14, +1.16]$ & $+0.17$$^{\dagger}$\;$[-0.17, +0.51]$ \\[2pt]
(h) Gap to oracle & $+5.84$$^{*}$\;$[+4.72, +6.93]$ & $+3.61$$^{*}$\;$[+2.79, +4.53]$ \\[2pt]
\bottomrule
\end{tabular*}
\tablenote{Units, signs and row contrasts follow Table~\ref{tab:matrix}; TOST margin: $\pm0.5$\pp. Brackets: 95\% re-planned intervals ($R=100$); $^{*}$ excludes zero; $^{\mathrm N}$ budget-normalised intervals ($R=2000$); $^{\dagger}$ regenerated with the original seed. $^{\S}$ Structural masses use fact, entity and relink actions; comparison rows use the full menu.}

\end{table*}

Tables~\ref{tab:matrix} and~\ref{tab:matrix-cross} compare representations at fixed actions and gains from adding actions, at budget $b=0.2$.

\begin{finding}[Development equivalence depends on the chosen margins]
\label{f:joint}
With action sets fixed and development margins chosen after inspecting the initial intervals, the lattice and independent-block models satisfy the TOST criteria in all five single-document configurations ($\pm1\%$) and both cross-document graphs ($\pm0.5$\pp{}). Equivalence with direct utility regression holds in four of five single-document comparisons and both cross-document graphs; the Claude comparison is inconclusive. All five comparisons with the full 8-class mask model pass. With LR, two of four cross-document comparisons pass.
\end{finding}

Two observations help interpret these results. Structural mass is small: $\Dcl$ is at most four facts within documents and accounts for 1.1\% and 0.8\% of cross-document errors for Qwen and GPT-4.1. The scalar-preserving first-order planner consequently matches the lattice planner in 29 of 30 single-document comparisons, with a one-error difference in the remaining comparison. Learned plans also select few entity reviews within documents---3--5 per budget, compared with 28--45 for the oracle---so much of their performance comes from fact ranking.

The size of an equivalence margin matters in this regime. For GPT-4.1 and mini, the development margin is approximately the entire oracle gain from adding entity reviews.

\begin{finding}[The fan-out heuristic leaves more errors across documents]
\label{f:node}
Adding entity reviews ranked by node error $\times$ fan-out increases remaining errors relative to fact-only review by 3.13 [1.92, 4.35]\pp{} for Qwen and 3.98 [3.02, 4.87]\pp{} for GPT-4.1 (Table~\ref{tab:matrix-cross}, row c). At fixed actions, the lattice model removes 4.62 and 4.70\pp{} more errors than the node-marginal model (row d). This latter comparison is confirmed on held-out data. Within documents, adding the node queue significantly increases errors for Claude and mini, while the fixed-action lattice comparison is significant for all five configurations with LR and only Claude with GB.
\end{finding}

The accounting identity in Remark~\ref{rem:marginal} identifies an overcounting mechanism: a node review receives credit for incident facts that also require another repair. Such multi-block errors account for 28.2\% and 29.1\% of cross-document errors, mostly combinations of a wrong link and a wrong local decision, compared with at most 6.14\% within documents. At these masses, the worst-case bound $(1+2m_{\max})\Delta$ exceeds the total error count, so the size of the queue's disadvantage is assessed empirically. The $\Dcl$ bound instead concerns the scalar-preserving first-order planner.

\begin{finding}[Learned plans capture part of the shared-action value]
\label{f:shared}
Oracle plans benefit from shared actions for every extractor. Across documents, the learned lattice model gains 1.46 [0.87, 1.98]\pp{} for Qwen and 0.69 [0.28, 1.02]\pp{} for GPT-4.1, capturing 21\% and 17\% of the respective oracle gains (Table~\ref{tab:matrix-cross}, rows a--b). Against a GB fact-scalar model restricted to fact reviews, it gains 1.49 [0.90, 2.07] and 0.72 [0.33, 1.09]\pp{} (row s). Within-document gains are smaller; only the Qwen comparison against the fact-scalar model is significant, at 0.32\pp{} or 16 facts (Table~\ref{tab:matrix}).
\end{finding}

Relinking supplies most of the oracle saving, with little additional benefit from relabel and split. The $\Dcl$ values of 66 and 54 apply to the fact, entity and relink menu; full-menu gains also account for one relabel per graph that breaks a correct member. Increasing model capacity from LR to GB helps four of five single-document configurations, while the cross-document capacity differences are inconclusive.

\section{What an additional verification pass buys}
\label{sec:verif}

\begin{table*}[t]
\caption{Development verification with fact and entity actions (lenient, GB).}
\label{tab:expA}
\centering
\compacttablestyle
\setlength{\tabcolsep}{5pt}
\begin{tabular*}{\textwidth}{@{\extracolsep{\fill}}ll ccc@{}}
\toprule
Extraction (verifier) & Quantity & $b=0.1$ (gate) & $b=0.2$ (gate) & $b=0.3$ \\
\midrule
\multirow{4}{*}{\makecell[l]{Qwen extraction\\(verifier GPT-4.1-mini)}} & A0 repair yield & 99.8\% (867/869) & 97.6\% (1,697/1,739) & 94.8\% (2,472/2,608) \\
 & A0 fact-only gap & $+2$ {\footnotesize$[-10, +14]$} & $+42$ {\footnotesize$[+22, +65]$}$^{*}$ & $+136$ {\footnotesize$[+99, +171]$}$^{*}$ \\
 & A1$-$A0: errors removed & $-8$ {\footnotesize$[-23, +6]$} & $+6$ {\footnotesize$[-9, +21]$} & $+33$ {\footnotesize$[+11, +53]$}$^{*}$ \\
 & \quad share of fact-only gap & $-400$\% {\footnotesize$[-937, +63]$} & $+14$\% {\footnotesize$[-29, +40]$} & $+24$\% {\footnotesize$[+9, +38]$}$^{*}$ \\
\midrule
\multirow{4}{*}{\makecell[l]{GPT-4.1 extraction\\(verifier Qwen3.5-27B)}} & A0 repair yield & 99.1\% (836/844) & 98.2\% (1,658/1,688) & 96.8\% (2,450/2,532) \\
 & A0 fact-only gap & $+8$ {\footnotesize$[+0, +18]$} & $+30$ {\footnotesize$[+16, +42]$}$^{*}$ & $+82$ {\footnotesize$[+63, +111]$}$^{*}$ \\
 & A1$-$A0: errors removed & $+8$ {\footnotesize$[-1, +18]$} & $+27$ {\footnotesize$[+12, +37]$}$^{*}$ & $+71$ {\footnotesize$[+48, +95]$}$^{*}$ \\
 & \quad share of fact-only gap & $+100$\% {\footnotesize$[+22, +257]$}$^{*}$ & $+90$\% {\footnotesize$[+59, +113]$}$^{*}$ & $+87$\% {\footnotesize$[+73, +97]$}$^{*}$ \\
\bottomrule
\end{tabular*}
\tablenote{A0 uses extraction features; A1 adds a generic support check. Repair yield is errors removed per budget unit; the fact-only gap compares A0 with the clairvoyant fact-review plan. Brackets: 95\% re-planned bootstrap intervals ($R=100$); $^{*}$ excludes zero. Populations are in Table~\ref{tab:populations}.}

\end{table*}

We augment the inexpensive extraction features with a generic support check: the verifier reads the source document and candidate fact, answers yes or no, and supplies a token-level $P(\text{yes})$. GPT-4.1-mini verifies Qwen extractions, and Qwen3.5-27B verifies GPT-4.1 extractions. The baseline and augmented models, labelled A0 and A1 in the tables, use the same GB lattice representation, fact and entity actions, and MILP planner. Failed checks retain the baseline features. A development gate required verification to improve both extractors at each of the 10\% and 20\% review budgets. Section~\ref{sec:confirm} compares these results with a pipeline rerun used to fit the held-out models.

\begin{finding}[Better error ranking gives budget-dependent audit gains]
\label{f:gate}
Generic verification raises AUROC from 0.8826 to 0.8954 on Qwen and from 0.8793 to 0.9295 on GPT-4.1. At a 10\% budget, neither extractor shows a significant audit gain. At 20\%, GPT-4.1 gains 27 [12, 37] repairs, while Qwen remains inconclusive. At 30\%, both improve: Qwen by 33 [11, 53] and GPT-4.1 by 71 [48, 95] repairs. The original development gate is therefore unmet (Table~\ref{tab:expA}).
\end{finding}

The baseline is already close to the fact-only oracle at the smallest budget, removing 867 errors with 869 budget units on Qwen and 836 with 844 on GPT-4.1. These correspond to repair yields of 0.998 and 0.991. Because entity reviews can repair several facts, repair yield measures errors removed per unit cost and can exceed one. Precision@budget applies specifically to a fact-review queue. The fact-only GB scalar queues remove 859 and 831 errors with the same 869 and 844 review slots, giving precision@budget of 98.8\% and 98.5\% and selection gaps of 10 and 13 errors (Table~\ref{tab:labelfree}). The two- and eight-error gaps above compare plans containing entity actions with a fact-only oracle; they are reference gaps, not upper bounds for the full action menu.

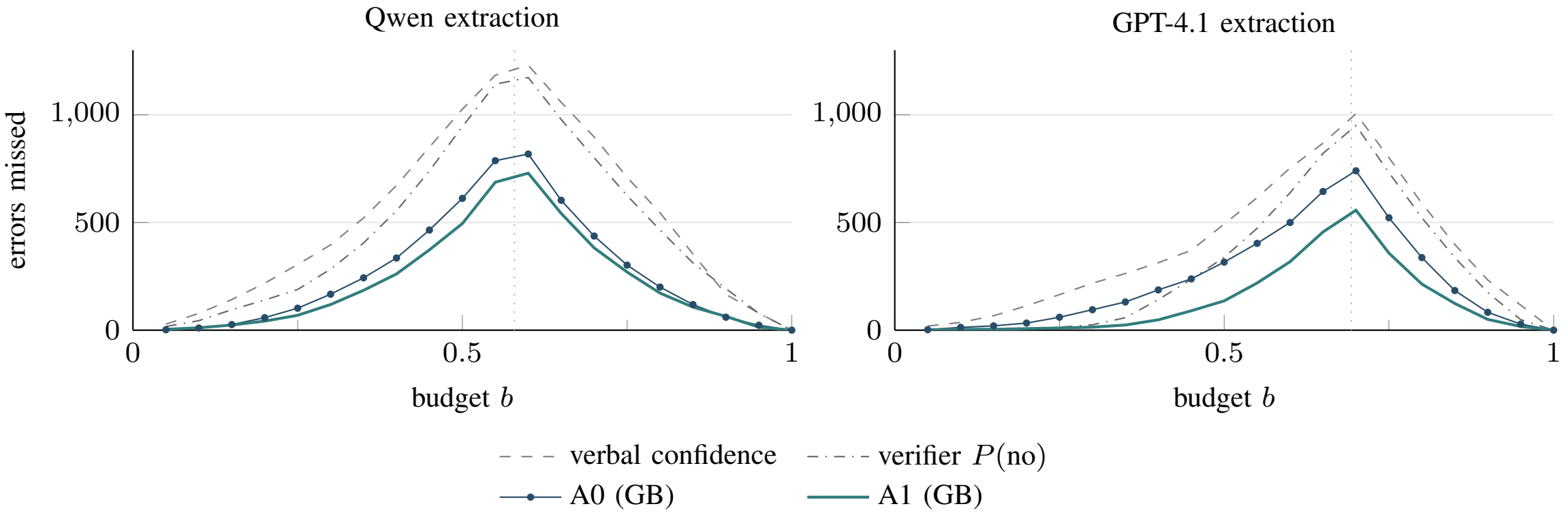
\begin{figure*}[t]
\centering
\begin{tikzpicture}
\begin{groupplot}[group style={group size=2 by 1, horizontal sep=1.0cm},
  width=0.44\textwidth, height=4.3cm, xmin=0, xmax=1, ymin=0, ymax=1300,
  xtick={0,0.5,1}, minor xtick={0.25,0.75}, xlabel={\footnotesize budget $b$}, tick label style={font=\footnotesize},
  ylabel style={font=\footnotesize}, title style={font=\footnotesize, yshift=-1ex},
  axis x line*=bottom, axis y line*=left, ymajorgrids, grid style={gray!20},
  legend style={font=\footnotesize, draw=none, legend columns=2,
    /tikz/every even column/.append style={column sep=6pt}}, legend cell align=left]
\nextgroupplot[title={Qwen extraction}, ylabel={errors missed}, legend to name=curveslegend]
\addplot[gray!60, dotted, forget plot] coordinates {(0.579,0) (0.579,1300)};
\addplot[gray, dashed, mark=none] coordinates {(0.05,28) (0.10,78) (0.15,142) (0.20,218) (0.25,305) (0.30,396) (0.35,521) (0.40,672) (0.45,850) (0.50,1026) (0.55,1184) (0.60,1230) (0.65,1058) (0.70,897) (0.75,710) (0.80,545) (0.85,354) (0.90,166) (0.95,79) (1.00,0)};
\addlegendentry{verbal confidence}
\addplot[gray!80!black, dash dot, mark=none] coordinates {(0.05,17) (0.10,44) (0.15,95) (0.20,140) (0.25,189) (0.30,283) (0.35,405) (0.40,554) (0.45,740) (0.50,945) (0.55,1142) (0.60,1173) (0.65,978) (0.70,801) (0.75,625) (0.80,465) (0.85,312) (0.90,193) (0.95,78) (1.00,0)};
\addlegendentry{verifier $P(\text{no})$}
\addplot[auditblue, mark=*, mark size=0.8pt] coordinates {(0.05,2) (0.10,10) (0.15,26) (0.20,58) (0.25,102) (0.30,167) (0.35,243) (0.40,335) (0.45,465) (0.50,612) (0.55,787) (0.60,818) (0.65,603) (0.70,437) (0.75,302) (0.80,200) (0.85,119) (0.90,60) (0.95,23) (1.00,0)};
\addlegendentry{A0 (GB)}
\addplot[auditteal, thick, mark=none] coordinates {(0.05,4) (0.10,11) (0.15,24) (0.20,42) (0.25,69) (0.30,119) (0.35,185) (0.40,261) (0.45,373) (0.50,495) (0.55,687) (0.60,729) (0.65,542) (0.70,382) (0.75,270) (0.80,172) (0.85,107) (0.90,64) (0.95,14) (1.00,0)};
\addlegendentry{A1 (GB)}
\nextgroupplot[title={GPT-4.1 extraction}]
\addplot[gray!60, dotted, forget plot] coordinates {(0.693,0) (0.693,1300)};
\addplot[gray, dashed, mark=none] coordinates {(0.05,19) (0.10,36) (0.15,67) (0.20,113) (0.25,165) (0.30,218) (0.35,263) (0.40,313) (0.45,371) (0.50,493) (0.55,615) (0.60,753) (0.65,868) (0.70,1005) (0.75,801) (0.80,589) (0.85,398) (0.90,233) (0.95,116) (1.00,0)};
\addplot[gray!80!black, dash dot, mark=none] coordinates {(0.05,2) (0.10,4) (0.15,6) (0.20,10) (0.25,14) (0.30,25) (0.35,58) (0.40,142) (0.45,236) (0.50,338) (0.55,473) (0.60,638) (0.65,821) (0.70,950) (0.75,732) (0.80,522) (0.85,337) (0.90,175) (0.95,49) (1.00,0)};
\addplot[auditblue, mark=*, mark size=0.8pt] coordinates {(0.05,2) (0.10,13) (0.15,20) (0.20,33) (0.25,60) (0.30,95) (0.35,131) (0.40,187) (0.45,238) (0.50,316) (0.55,403) (0.60,500) (0.65,644) (0.70,740) (0.75,522) (0.80,337) (0.85,184) (0.90,83) (0.95,28) (1.00,0)};
\addplot[auditteal, thick, mark=none] coordinates {(0.05,2) (0.10,4) (0.15,4) (0.20,7) (0.25,10) (0.30,14) (0.35,24) (0.40,48) (0.45,90) (0.50,136) (0.55,219) (0.60,318) (0.65,456) (0.70,557) (0.75,358) (0.80,214) (0.85,124) (0.90,50) (0.95,17) (1.00,0)};
\end{groupplot}
\end{tikzpicture}\\[1pt]
\pgfplotslegendfromname{curveslegend}
\caption{Errors that a fact-review ranking misses relative to the clairvoyant fact-review plan, over the full budget range (fact-only top-$b$ plans). Dotted line: the share of wrong facts. A0 and A1 are GB models fitted on dev labels; the other two rankers use no labels.}
\label{fig:curves}
\end{figure*}

In the fact-only sweep, headroom peaks near the share of wrong facts, 0.58 and 0.69 for Qwen and GPT-4.1 (Figure~\ref{fig:curves}). At a 50\% budget, the baseline misses 612 and 316 repairs available to the oracle; verification reduces these gaps to 495 and 136.

\begin{table*}[t]
\caption{Label-free and label-trained fact-review plans (dev, lenient).}
\label{tab:labelfree}
\centering
\tablestyle
\setlength{\tabcolsep}{6pt}
\begin{tabular*}{\textwidth}{@{\extracolsep{\fill}}l c ccc@{}}
\toprule
Ranker & AUROC & $b=0.1$ & $b=0.2$ & $b=0.3$ \\
\midrule
\multicolumn{5}{@{}l}{\textbf{Qwen extraction} (verifier GPT-4.1-mini): benchmark errors removed by the top $\lfloor bN\rfloor$ facts}\\
Verbal confidence (label-free) & 0.754 & 791 & 1,521 & 2,212 \\
Triple log-probability (label-free) & 0.631 & 758 & 1,417 & 1,992 \\
Verifier $P(\text{no})$ alone (label-free) & 0.783 & 825 & 1,599 & 2,325 \\
A0: GB, fact scalar & 0.873 & 859 & 1,681 & 2,441 \\
A1: GB, fact scalar & 0.892 & 858 & 1,697 & 2,489 \\
A0: GB, lattice & 0.883 & 859 & 1,688 & 2,466 \\
A1: GB, lattice & 0.895 & 858 & 1,702 & 2,502 \\
Clairvoyant fact-only plan & -- & 869 & 1,739 & 2,608 \\
\multicolumn{5}{@{}l}{\textit{Differences in errors removed (re-planned 95\%, 1,000 resamples)}}\\
A0 $-$ verbal confidence & & $+68$ {\footnotesize$[+42, +101]$}$^{*}$ & $+160$ {\footnotesize$[+121, +201]$}$^{*}$ & $+229$ {\footnotesize$[+172, +281]$}$^{*}$ \\
A0 $-$ log-probability & & $+101$ {\footnotesize$[+81, +123]$}$^{*}$ & $+264$ {\footnotesize$[+226, +302]$}$^{*}$ & $+449$ {\footnotesize$[+394, +506]$}$^{*}$ \\
Verifier $P(\text{no})$ $-$ verbal & & $+34$ {\footnotesize$[+2, +66]$}$^{*}$ & $+78$ {\footnotesize$[+29, +122]$}$^{*}$ & $+113$ {\footnotesize$[+46, +169]$}$^{*}$ \\
A1 $-$ verifier $P(\text{no})$ & & $+33$ {\footnotesize$[+18, +53]$}$^{*}$ & $+98$ {\footnotesize$[+71, +129]$}$^{*}$ & $+164$ {\footnotesize$[+130, +208]$}$^{*}$ \\
A1 $-$ A0 (fact scalar) & & $-1$ {\footnotesize$[-8, +5]$} & $+16$ {\footnotesize$[-2, +29]$} & $+48$ {\footnotesize$[+27, +71]$}$^{*}$ \\
Clairvoyant $-$ A0 (fact scalar) & & $+10$ {\footnotesize$[+5, +17]$}$^{*}$ & $+58$ {\footnotesize$[+41, +77]$}$^{*}$ & $+167$ {\footnotesize$[+132, +205]$}$^{*}$ \\
Clairvoyant $-$ verbal confidence & & $+78$ {\footnotesize$[+52, +110]$}$^{*}$ & $+218$ {\footnotesize$[+173, +263]$}$^{*}$ & $+396$ {\footnotesize$[+333, +456]$}$^{*}$ \\
\midrule
\multicolumn{5}{@{}l}{\textbf{GPT-4.1 extraction} (verifier Qwen3.5-27B): benchmark errors removed by the top $\lfloor bN\rfloor$ facts}\\
Verbal confidence (label-free) & 0.792 & 808 & 1,575 & 2,314 \\
Triple log-probability (label-free) & 0.631 & 772 & 1,476 & 2,144 \\
Verifier $P(\text{no})$ alone (label-free) & 0.846 & 840 & 1,678 & 2,507 \\
A0: GB, fact scalar & 0.874 & 831 & 1,655 & 2,437 \\
A1: GB, fact scalar & 0.926 & 840 & 1,681 & 2,518 \\
A0: GB, lattice & 0.879 & 835 & 1,657 & 2,449 \\
A1: GB, lattice & 0.929 & 842 & 1,683 & 2,519 \\
Clairvoyant fact-only plan & -- & 844 & 1,688 & 2,532 \\
\multicolumn{5}{@{}l}{\textit{Differences in errors removed (re-planned 95\%, 1,000 resamples)}}\\
A0 $-$ verbal confidence & & $+23$ {\footnotesize$[+8, +43]$}$^{*}$ & $+80$ {\footnotesize$[+54, +103]$}$^{*}$ & $+123$ {\footnotesize$[+90, +152]$}$^{*}$ \\
A0 $-$ log-probability & & $+59$ {\footnotesize$[+43, +79]$}$^{*}$ & $+179$ {\footnotesize$[+148, +208]$}$^{*}$ & $+293$ {\footnotesize$[+255, +331]$}$^{*}$ \\
Verifier $P(\text{no})$ $-$ verbal & & $+32$ {\footnotesize$[+17, +49]$}$^{*}$ & $+103$ {\footnotesize$[+78, +127]$}$^{*}$ & $+193$ {\footnotesize$[+155, +224]$}$^{*}$ \\
A1 $-$ verifier $P(\text{no})$ & & $+0$ {\footnotesize$[-3, +4]$} & $+3$ {\footnotesize$[-1, +8]$} & $+11$ {\footnotesize$[+2, +23]$}$^{*}$ \\
A1 $-$ A0 (fact scalar) & & $+9$ {\footnotesize$[+0, +17]$} & $+26$ {\footnotesize$[+15, +41]$}$^{*}$ & $+81$ {\footnotesize$[+56, +104]$}$^{*}$ \\
Clairvoyant $-$ A0 (fact scalar) & & $+13$ {\footnotesize$[+5, +21]$}$^{*}$ & $+33$ {\footnotesize$[+23, +47]$}$^{*}$ & $+95$ {\footnotesize$[+71, +117]$}$^{*}$ \\
Clairvoyant $-$ verbal confidence & & $+36$ {\footnotesize$[+22, +53]$}$^{*}$ & $+113$ {\footnotesize$[+88, +136]$}$^{*}$ & $+218$ {\footnotesize$[+181, +248]$}$^{*}$ \\
\bottomrule
\end{tabular*}
\tablenote{Rankers select the riskiest $\lfloor bN\rfloor$ facts, breaking ties by fact-key hash. Triple log-probability sums head, relation and tail tokens; verifier scores replace 25 failed Qwen-extraction calls with the median. A0/A1 use out-of-fold GB predictions trained on dev labels. Intervals reselect the top-$b$ set in each document-cluster resample; $^{*}$ excludes zero.}

\end{table*}

\begin{finding}[Verification improves on label-free confidence rankers]
\label{f:labelfree}
The verifier's $P(\text{no})$ outperforms the extractor's verbal confidence at each tested budget. It adds 34 [2, 66], 78 and 113 repairs for Qwen at budgets 10\%, 20\% and 30\%, and 32 [17, 49], 103 and 193 for GPT-4.1 (95\% intervals at all three budgets appear in Table~\ref{tab:labelfree}). On GPT-4.1 it approaches the learned augmented ranker, whose additional gains are 0 [-3, 4], 3 [-1, 8] and 11 [2, 23].
\end{finding}

The baseline matters: the label-trained extraction model already removes 68, 160 and 229 more errors than verbal confidence on Qwen, and 23, 80 and 123 more on GPT-4.1. In a sensitivity analysis that increases precision to 0.8 by subsampling wrong facts, the 10\%-budget verification gain rises from $-1$ to $22.2$ repairs for Qwen and from $9$ to $56.0$ for GPT-4.1, using unchanged fitted scores.

Verification also changes many decisions whose effects cancel. At a 20\% budget on GPT-4.1, 1,355 review decisions change: 670 beneficial and 643 harmful changes account for the net gain of 27.

\subsection{Control checks, cost and repair outcome}
We compare G with targeted relation (R) and evidence-first (S) menus constructed from development statistics. Repeated Qwen verifier calls return identical answers, so a reworded second generic call (G2p) provides a prompt-variation control. R and S do not improve on G: at $b=0.2$, R+S changes repairs by $+2$ for Qwen and $-4$ for GPT-4.1; at $b=0.3$ the changes are $-8$ and $-5$. R+S uses 1.22 times the tokens of G+G2p.

To examine verifier dependence, a cross-family \texttt{kimi-k2.5} check on the first 100 documents gives an extraction-dependent AUROC gap of $+0.055$ [$+0.028$, $+0.083$], compared with $+0.041$ [$+0.013$, $+0.070$] for the default verifiers; the difference in differences is $-0.014$ [$-0.043$, $+0.010$].

Generic verification uses 623 tokens per fact for Qwen extractions and 672 for GPT-4.1 extractions. Under the experimental review prices, its GPT-4.1 gain at $b=0.2$ is equivalent to about 28 extra fact reviews, or about 199k verification tokens per saved review; at $b=0.3$, the corresponding figures are about 82 reviews and 69k tokens.

The repair objective counts both correction and deletion. Under a model that corrects a fact when a gold target exists and deletes it otherwise, GPT-4.1 verification at $b=0.2$ fixes 1,685 errors versus 1,658 for the baseline, but corrects only 288 versus 345. KG precision changes by $-0.32$\pp{} [$-0.64$, $-0.03$]. On GPT-4.1, G, G+G2p and R+S at 10\% and 20\% yield six negative precision changes; four unadjusted intervals exclude zero. Residual-error reduction and the precision of the retained graph are different measurements.

\section{Held-out confirmation}
\label{sec:confirm}

We froze the protocol on 2026-10-17 before held-out alignment, using Re-DocRED \textbf{test} (500 documents) for confirmation and \textbf{train500} (500 hash-selected training documents) for replication. Models fitted on dev were applied unchanged. Each dataset has ten primary model--budget comparisons, evaluated with 1,000 re-planned document-cluster bootstrap samples and Holm correction. A hypothesis is confirmed only when all its specified comparisons pass.

The protocol tests three hypotheses, labelled H1--H3 in Table~\ref{tab:confirm}. The \emph{equivalence hypothesis} compares joint and independent-block models at a 20\% budget, with margins of $\pm0.5\%$ of wrong facts within documents and $\pm0.25$\pp{} across documents. The \emph{verification hypothesis} requires generic verification to improve both extractors at budgets of 20\% and 30\%. The \emph{fan-out hypothesis} compares the node-marginal and joint models across documents at a 20\% budget with identical action sets.

\begin{table*}[t]
\caption{Held-out results under the frozen protocol (lenient, GB).}
\label{tab:confirm}
\centering
\tablestyle
\begin{tabular*}{\textwidth}{@{\extracolsep{\fill}}lll@{}}
\toprule
Extractor / budget & \textbf{test} & \textbf{train500} \\
\midrule
\addlinespace[3pt]
\multicolumn{3}{@{}l}{\tablegroup{H1, within documents: lattice $-$ independent (errors; TOST, $\pm0.5\%$ of wrong facts)}} \\
\quad Qwen & $+19$ [$+6$, $+35$] & $+5$ [$-6$, $+12$] $\checkmark$ \\
\quad GPT-4.1 & $-4$ [$-14$, $+6$] $\checkmark$ & $+3$ [$-6$, $+13$] $\checkmark$ \\
\addlinespace[3pt]
\multicolumn{3}{@{}l}{\tablegroup{H1, across documents: joint $-$ independent (\pp; TOST, $\pm0.25$\pp)}} \\
\quad Qwen & $+0.31$ [$-0.05$, $+0.72$] & $+0.32$ [$+0.08$, $+0.60$] \\
\quad GPT-4.1 & $+0.27$ [$+0.08$, $+0.48$] & $+0.06$ [$-0.11$, $+0.25$] \\
\addlinespace[3pt]
\multicolumn{3}{@{}l}{\tablegroup{H2, verification: A0 $-$ A1 (errors; one-sided)}} \\
\quad Qwen, $b=0.2$ & $+6$ [$-11$, $+25$] & $+10$ [$-1$, $+21$] \\
\quad Qwen, $b=0.3$ & $+40$ [$+12$, $+72$] $\checkmark$ & $+22$ [$+7$, $+40$] $\checkmark$ \\
\quad GPT-4.1, $b=0.2$ & $+32$ [$+15$, $+53$] $\checkmark$ & $+31$ [$+20$, $+45$] $\checkmark$ \\
\quad GPT-4.1, $b=0.3$ & $+59$ [$+32$, $+95$] $\checkmark$ & $+60$ [$+40$, $+79$] $\checkmark$ \\
\addlinespace[3pt]
\multicolumn{3}{@{}l}{\tablegroup{H3, across documents: joint $-$ node marginal (\pp; one-sided)}} \\
\quad Qwen & $-4.28$ [$-5.01$, $-3.35$] $\checkmark$ & $-3.98$ [$-4.79$, $-3.16$] $\checkmark$ \\
\quad GPT-4.1 & $-4.40$ [$-5.27$, $-3.55$] $\checkmark$ & $-4.74$ [$-5.44$, $-3.93$] $\checkmark$ \\
\midrule
Verdict H1 & not confirmed (1/4) & not replicated (2/4) \\
Verdict H2 & not confirmed (3/4) & not replicated (3/4) \\
Verdict H3 & \textbf{confirmed} (2/2) & \textbf{replicated} (2/2) \\
\bottomrule
\end{tabular*}
\tablenote{Positive means the first plan leaves more errors. Intervals are unadjusted re-planned document-cluster bootstrap intervals ($R=1000$): 90\% for H1, 95\% otherwise. $\checkmark$ denotes Holm rejection across ten comparisons per split ($\alpha=0.05$). Each hypothesis requires all its comparisons to pass. Models were fitted on dev only; $b=0.2$ unless noted.}

\end{table*}

\begin{finding}[The fan-out disadvantage replicates on held-out data]
\label{f:confirm-node}
The joint model leaves 4.28 and 4.40\pp{} fewer errors than the node-marginal model on test, and 3.98 and 4.74\pp{} fewer on train500, for Qwen and GPT-4.1 respectively. All four comparisons pass Holm correction, confirming and replicating the hypothesis. The joint planner's separate gain from opening shared actions, relative to its own fact-only plan, is 0.89--1.43\pp{} in descriptive point estimates (Table~\ref{tab:cross-context}).
\end{finding}

Table~\ref{tab:cross-context} places independent blocks alongside the other plans. Relative to the common fact-scalar reference, node marginals leave 2.52--3.45\pp{} more errors, independent blocks leave 1.22--1.79\pp{} fewer, and the joint model leaves 0.91--1.46\pp{} fewer. Opening shared actions relative to each model's own fact-only plan gives independent blocks 70/68 additional repairs on test and 109/81 on train500 (Qwen/GPT-4.1); the joint gains are 60/59 and 88/82. Oracle gains are 387/296 and 366/307. The factorised model captures shared-action value without a joint training target; the frozen equivalence tests below assess the stronger claim of comparable performance.

\begin{finding}[Verification gains recur in three of four configurations]
\label{f:confirm-verif}
GPT-4.1 gains 32 and 59 repairs at the 20\% and 30\% budgets on test, and 31 and 60 on train500. Qwen gains 40 and 22 repairs at the 30\% budget on the two datasets. All six comparisons pass Holm correction. Qwen's 20\%-budget gains are 6 [-11, 25] and 10 [-1, 21], so the verification hypothesis remains unconfirmed under the all-comparisons rule.
\end{finding}

The same Qwen configuration falls short on development, test and train500. Its fact-only reference gap is larger than GPT-4.1's on both held-out datasets, which rules out smaller headroom as a complete explanation of this pattern. Section~\ref{sec:practical} examines this comparison in more detail.

\begin{finding}[The tighter equivalence criterion remains inconclusive]
\label{f:confirm-joint}
One of four equivalence comparisons passes on test and two of four on train500. The three unadjusted 90\% intervals that exclude zero all favour independent blocks: the lattice model leaves 19 [6, 35] more errors for within-document Qwen on test; the joint model leaves 0.27 [0.08, 0.48]\pp{} more for cross-document GPT-4.1 on test and 0.32 [0.08, 0.60]\pp{} more for cross-document Qwen on train500. No reported interval excludes zero in favour of the joint model.
\end{finding}

\paragraph{Sensitivity to linking and refitting}
Before freezing the held-out models, we regenerated cross-document features with fresh linking and Wikipedia resolution, then refitted the models. Qwen's joint-minus-independent contrast shifted from $-0.05$ to $+0.28$\pp{}, a 0.33\pp{} change exceeding the fixed $\pm0.25$\pp{} equivalence margin. The regenerated GPT-4.1 interval, [$+0.02$, $+0.46$]\pp{}, also crossed that margin. Held-out models use the regenerated features; the protocol retained the original margins.

\paragraph{Within-document label sensitivity}
Single-document $\Dcl$ is one or two facts under lenient labels and 7--15 under strict labels. At $b=0.2$, adding entity actions gives the learned plans $-3$ to $+6$ additional repairs under lenient alignment, compared with 67--101 for the oracle. In the pre-specified strict-label sensitivity analysis, learned gains range from $+11$ to $+59$, with oracle gains of 165--303 (Table~\ref{tab:confirm2}).

\begin{table}[t]
\caption{Four cross-document plans on held-out data ($b=0.2$).}
\label{tab:cross-context}
\centering\compacttablestyle
\setlength{\tabcolsep}{2.5pt}
\begin{tabular*}{\columnwidth}{@{\extracolsep{\fill}}llrrrrr@{}}
\toprule
Set & Extractor & F & N & I & J & Gain I/J \\
\midrule
test & Qwen & 4057 & 4251 & 3987 & 4005 & 70/60 \\
test & GPT-4.1 & 4908 & 5137 & 4827 & 4845 & 68/59 \\
train500 & Qwen & 4453 & 4608 & 4343 & 4363 & 109/88 \\
train500 & GPT-4.1 & 4987 & 5216 & 4896 & 4900 & 81/82 \\
\bottomrule\end{tabular*}
\tablenote{Remaining benchmark errors; lower is better (lenient labels, GB). F: fact scalar, fact reviews only. N: node marginals; I: independent blocks; J: joint model, all with the full shared-action menu. Gain I/J compares each model with its own fact-only plan. These are descriptive point estimates; frozen H1/H3 intervals and decisions are in Table~\ref{tab:confirm}.}
\end{table}

\begin{table}[t]
\caption{Within-document secondary results (point estimates in errors).}
\label{tab:confirm2}
\centering
\compacttablestyle
\setlength{\tabcolsep}{2.5pt}
\begin{tabular*}{\columnwidth}{@{\extracolsep{\fill}}llrrrr@{}}
\toprule
 & & \multicolumn{2}{c}{test} & \multicolumn{2}{c}{train500} \\
 & & Qwen & GPT & Qwen & GPT \\
\midrule
H2, $b=0.1$ & lenient & $-2$ & $+3$ & $+2$ & $+9$ \\
H2, $b=0.5$ & lenient & $+50$ & $+124$ & $+41$ & $+109$ \\
H2, $b=0.2$ & strict & $-6$ & $+25$ & $+32$ & $+26$ \\
H2, $b=0.3$ & strict & $+9$ & $+57$ & $+51$ & $+46$ \\
H1 within & strict & $+11$ & $-7$ & $+5$ & $-11$ \\
\midrule
Entity value, learned & lenient & $+2$ & $+0$ & $+6$ & $-3$ \\
Entity value, oracle & lenient & $+97$ & $+67$ & $+101$ & $+95$ \\
Entity value, learned & strict & $+59$ & $+36$ & $+58$ & $+11$ \\
Entity value, oracle & strict & $+303$ & $+189$ & $+271$ & $+165$ \\
\bottomrule
\end{tabular*}
\tablenote{$b=0.2$ unless noted. H1: lattice minus independent residual error; H2: baseline minus verified residual error. Entity value: additional repairs from entity reviews for the GB lattice planner or oracle.}

\end{table}
\subsection{Held-out design and inference}
Train500 consists of the 500 training documents with the smallest SHA-256 title hashes after removing overlaps with dev and test. Those documents had previously contributed aggregate relation frequencies to a recall screen; these frequencies were not used by the audit models, features or thresholds. We therefore call this set a replication. Extraction failures count as misses: four Qwen and one GPT-4.1 documents on test, none on train500.

The 20\% and 30\% verification budgets were selected from development fact-only reference gaps near or above 30 errors. Before freezing, we repeated generic verification and refitted on regenerated features. The gains of 27, 33 and 71 in Section~\ref{sec:verif} became 25, 28 and 68, while GPT-4.1's 20\% reference gap fell from 30 to 29; both budgets remained in the protocol. Held-out verification models use these regenerated features.

Bootstrap $p$-values use $(1+\mathrm{count})/(1000+1)$, with one Holm family of ten comparisons per dataset; within-document seed-1--4 refits reproduce seed-0 point estimates. Aligned facts, transferred probabilities, document-level residuals and bootstrap arrays are included in the reproduction package.

\section{From confidence scores to useful audit decisions}
\label{sec:practical}

\begin{table}[t]
\caption{Headroom and verification gains, in benchmark errors.}
\label{tab:headroom}
\centering\compacttablestyle
\setlength{\tabcolsep}{3pt}
\begin{tabular*}{\columnwidth}{@{\extracolsep{\fill}}llrrrr@{}}
\toprule
Set & Extractor & $b$ & $D_{\rm fact}$ & $D_{\rm full}$ & Gain \\
\midrule
test & Qwen & 0.2 & 58 & 155 & $+6$ \\
 & Qwen & 0.3 & 189 & 286 & $+40\checkmark$ \\
 & GPT-4.1 & 0.2 & 44 & 111 & $+32\checkmark$ \\
 & GPT-4.1 & 0.3 & 107 & 174 & $+59\checkmark$ \\
\midrule
train500 & Qwen & 0.2 & 40 & 141 & $+10$ \\
 & Qwen & 0.3 & 102 & 203 & $+22\checkmark$ \\
 & GPT-4.1 & 0.2 & 37 & 132 & $+31\checkmark$ \\
 & GPT-4.1 & 0.3 & 79 & 174 & $+60\checkmark$ \\
\bottomrule
\end{tabular*}
\tablenote{$D_{\rm fact}=R_{\rm A0}-R^*_{\rm fact}$; $D_{\rm full}=R_{\rm A0}-R^*_{\rm full}$ uses the baseline's fact+entity actions. Gain: $R_{\rm A0}-R_{\rm A1}$. Intervals and Holm decisions ($\checkmark$) are in Table~\ref{tab:confirm}.}

\end{table}

\paragraph{Turning headroom into gains}
Table~\ref{tab:headroom} compares the verification gains with two oracle references. At the 20\% budget, Qwen's $D_{\rm fact}$ is 58 on test and 40 on train500, yet verification gains are only $+6$ and $+10$ repairs. GPT-4.1 has smaller gaps of 44 and 37, with gains of $+32$ and $+31$, respectively. Headroom measures the available repairs; a verification score must change which facts are selected to realise them. For fact-review plans, the following one-step identity characterises this requirement through score crossings at the selection boundary.

For unit-cost fact reviews, let $r_i$ be the true error posterior and let $\tau,\tau'$ be the $B$-th and $(B+1)$-th largest values. If one diagnostic updates only $r_i$ to $r_i^O$, its value is $\E[(\tau'-r_i^O)_+]$ for a selected fact and $\E[(r_i^O-\tau)_+]$ for an unselected fact: exactly one fact is exchanged when its updated score crosses the opposite cutoff. This is the one-step expected value of sample information for the selection decision \citep{strong-etal-2015-estimating}.

\paragraph{Sensitivity to failed repairs}
We evaluate the four policies in Table~\ref{tab:cross-context} under hypothetical execution failures, holding their plans fixed. Each attempted action independently succeeds or leaves the graph unchanged, with its cost charged in either case. Fact and shared actions have separate success probabilities, each in $\{1,0.85,0.70\}$, and successful relabels retain their recorded damage to correct members. Exact expectations enumerate the relevant action outcomes for each fact, recovering all 16 ideal residual counts when both success probabilities equal one. Across all four configurations and nine probability pairs, independent and joint planning each leave fewer expected errors than both baselines. Their smallest gains over fact-only review are 0.39 and 0.24\pp{}, respectively.

\section{Label validity and extraction coverage}
\label{sec:bench}

\begin{table}[t]
\caption{LLM adjudication of benchmark discrepancies.}
\label{tab:artefacts}
\centering\compacttablestyle
\setlength{\tabcolsep}{2.5pt}
\begin{tabular*}{\columnwidth}{@{\extracolsep{\fill}}lcc@{}}
\toprule
Quantity & Qwen & GPT-4.1 \\
\midrule
\makecell[l]{Possible gold\\omissions} & \makecell{0.205\\$[0.161,0.248]$} & \makecell{0.158\\$[0.121,0.197]$} \\[2pt]
\makecell[l]{Convention\\mismatches} & \makecell{0.227\\$[0.192,0.263]$} & \makecell{0.202\\$[0.169,0.239]$} \\[2pt]
\makecell[l]{Relation judgement\\agreement ($\kappa$)} & \makecell{0.451\\$[0.243,0.670]$} & \makecell{0.390\\$[0.201,0.576]$} \\[2pt]
\makecell[l]{Conflicting KB link:\\judged gold error} & \makecell{0.650\\$[0.560,0.733]$} & \makecell{0.633\\$[0.549,0.716]$} \\[2pt]
\makecell[l]{Precision if omissions\\are accepted} & $0.421\to0.54$ & $0.307\to0.42$ \\
\bottomrule
\end{tabular*}
\tablenote{The first two rows are share-weighted means of two cross-family judges. Convention mismatches include extractor errors. Brackets: unadjusted 95\% bootstrap intervals. Sample sizes and reference-judge comparisons are in the text.}
\end{table}

Benchmark labels determine which repairs count as successful. Two judges from families other than the extractors, \nolinkurl{claude-opus-5-5} and \nolinkurl{gemini-3.1-pro-preview}, adjudicate stratified samples of 150 spurious facts, 50 relation errors, 50 argument errors and 170 links per extractor. \nolinkurl{gpt-6-astra} with \texttt{xhigh} reasoning provides reference judgements (Table~\ref{tab:artefacts}). Cohen's $\kappa$ measures agreement among these model judges; the labels have no human validation.

\begin{finding}[Judges identify possible omissions and convention mismatches]
\label{f:artefacts}
Share-weighted estimates from the two judges' stratified samples classify 20.5\% of Qwen's and 15.8\% of GPT-4.1's benchmark-wrong facts as supported by the document but absent from the gold labels. Treating these judgements as correct in a sensitivity calculation raises precision from 0.421 to 0.54 and from 0.307 to 0.42; the reference judge gives 0.544 and 0.403. A further 22.7\% and 20.2\% are classified as convention mismatches.
\end{finding}

Many convention mismatches concern relation labels, where judge agreement is moderate ($\kappa=0.451$ and $0.390$). Some reflect dataset-specific usage, while others remain errors under Wikidata's definitions. For example, 30 of 48 GPT-4.1 P276 errors use a location relation where the gold uses the administrative-entity relation P131. Combining convention mismatches with possible omissions therefore does not estimate a gold-error rate.

\begin{finding}[Judges flag many conflicting gold links as incorrect]
\label{f:links}
When prediction and gold point to different KB entities, the judges classify the gold link as incorrect in 65.0\% [56.0, 73.3] of sampled Qwen cases and 63.3\% [54.9, 71.6] of GPT-4.1 cases ($\kappa=0.87$ and $0.91$). The reference judge gives 64.2\% and 66.6\%.
\end{finding}

\paragraph{Extraction coverage}
The five configurations recover 1,485 (mini), 2,560 (GPT-4.1), 3,461 (Qwen-R), 3,600 (Qwen) and 3,860 (Claude) of the 17,284 development gold facts. Their union covers 6,131 facts, or 35.47\%.

Recovering additional facts requires extraction-side actions. In our development screens, deterministic rule compilation adds 193 benchmark-correct facts for GPT-4.1 and 245 for Qwen, alongside additional incorrect and unaligned predictions. In 300 additional calls on 50 documents, generic re-extraction adds 54 benchmark-correct facts across both extractors, random source windows add 15 and low-coverage windows add five under strict scoring. Low-coverage minus random is $-0.10$ [$-0.29$, $0.02$] facts per document and extractor.

\subsection{Alignment and annotation sensitivity}
Across documents, correcting links that both judges call clear gold errors changes the oracle cost saving at 50\% residual error from 13.75\% to 9.80\% [9.19, 11.00] for Qwen and from 8.12\% to 5.85\% [5.11, 6.56] for GPT-4.1. These calculations impute unadjudicated links from stratum posteriors. The learned-plan comparisons and structural masses retain the original link labels. Within documents, the lenient merge heuristic agrees with the reference judge on 77\% and 78\% of merged nodes; the learned entity comparisons were not recomputed from these judgments.

Re$^2$-DocRED \citep{heng-etal-2026-re2docred} adds 4,719 unique development triples, all on entity pairs that already had a gold relation. These additions overlap none of the 2,467 Qwen or 3,014 GPT-4.1 spurious facts in our check, but the span-aligned sensitivity check accepts 50 and 36 facts previously classified as relation or relation-and-direction errors. Precision consequently changes from 0.4205 to 0.4263 and from 0.3069 to 0.3111. This check evaluates label sensitivity with the existing alignment; audit-plan rankings were not recomputed.

\subsection{Structural screening and DWIE policy comparison}
\label{sec:development}

We screen SciREX and DWIE for structural mass beyond the DocRED family, using 66 SciREX dev documents, 306 SciREX train documents and a shared 300-document DWIE training subset selected by SHA-256 of document IDs \citep{jain-etal-2020-scirex,zaporojets-etal-2021-dwie}. We evaluate the first extraction attempt per document, counting unparsed outputs as misses (three GPT-4.1 and one Qwen document on DWIE). The gate for extending held-out evaluation required both extractors to have at least 30 $\Dcl$ facts and at least 2\% of errors within documents or 1\% across documents.

SciREX has 74.4--85.0\% unverifiable predictions and $\Dcl$ of zero to four facts; its four-argument records did not include an audited numerical score. DWIE within-document $\Dcl$ is zero for both extractors. Across documents, it is 20/2,995 (0.67\%) for GPT-4.1 and 27/2,834 (0.95\%) for Qwen. Strict labels give 50/3,328 (1.50\%) and 90/3,359 (2.68\%), while eligibility uses lenient labels. Thus neither dataset passes the gate for held-out extension.

\begin{table}[t]
\caption{DWIE audit-policy comparison on 300 training documents.}
\label{tab:development}
\centering\compacttablestyle
\setlength{\tabcolsep}{2.5pt}
\begin{tabular*}{\columnwidth}{@{\extracolsep{\fill}}lrrrrr@{}}
\toprule
Extractor & $b$ & F & N & I & J \\
\midrule
Qwen & 0.1 & 2317 & 2324 & 2300 & 2303 \\
Qwen & 0.2 & 1831 & 1843 & 1816 & 1818 \\
Qwen & 0.3 & 1393 & 1409 & 1368 & 1389 \\
\midrule
GPT-4.1 & 0.1 & 2478 & 2524 & 2485 & 2481 \\
GPT-4.1 & 0.2 & 1998 & 2043 & 2003 & 1997 \\
GPT-4.1 & 0.3 & 1563 & 1604 & 1552 & 1549 \\
\bottomrule
\end{tabular*}
\tablenote{Remaining benchmark errors; definitions follow Table~\ref{tab:cross-context}. Lenient labels, GB, five document-grouped folds.}
\end{table}

A post-hoc comparison on the same DWIE training sample evaluates the four policies at low $\Dcl$ (Table~\ref{tab:development}). Using the saved extractions, alignments and resolved links, it fits the existing GB representations in five document-grouped folds, with seed 0 and the same hyperparameters and review costs, at budgets 10\%, 20\% and 30\%. Cross-document evaluation uses 2018 Wikipedia link labels with unvalidated resolution differences; full-menu accounting includes one GPT-4.1 relabel that can break a correct member.

At 20\%, the marginal-risk plan leaves 12 more errors than fact-only review for Qwen and 45 more for GPT-4.1. Independent blocks leave 15 fewer for Qwen but five more for GPT-4.1; the joint gains are 13 and one. Independent-minus-node contrasts are $-0.95$ [$-2.31$, $+0.62$]\pp{} and $-1.34$ [$-2.31$, $-0.43$]\pp{}, respectively, with unadjusted 95\% budget-normalised document-bootstrap intervals (2,000 resamples, fitted scores and plans fixed). Shared actions improve each independent model over its own fact-only plan by 18 and three repairs, so gains against a common scalar baseline also depend on the fact ranking.

\section{Related work}
\label{sec:related}

\paragraph{Human-in-the-loop KG quality}
KG quality estimation uses observations of selected facts to assess a graph \citep{ojha-talukdar-2017-kgeval,gao-etal-2019-efficient,marchesin-silvello-2024-efficient,qi-etal-2022-evaluating}. Joint question selection in entity resolution and uncertainty reduction in schema matching exploit dependencies between decisions \citep{whang-etal-2013-question,zhang-etal-2013-reducing}. Interactive extraction with constrained CRFs \citep{kristjansson-etal-2004-interactive} and probabilistic repair with HoloClean \citep{rekatsinas-etal-2017-holoclean} supply related intervention settings.

\paragraph{Confidence estimation and calibration}
Verbal confidence estimates \citep{tian-etal-2023-just,xiong-etal-2024-can}, self-evaluation \citep{kadavath-etal-2022-language}, sampling-based consistency \citep{manakul-etal-2023-selfcheckgpt,kuhn-etal-2023-semantic} and token-level uncertainty \citep{fadeeva-etal-2024-fact} provide different signals of model reliability. Extraction-oriented methods combine these with features of documents and structured outputs \citep{goh-mueller-2026-construct,kumar-2026-extractconf}.

ReDD provides calibrated relation extraction \citep{chao-etal-2025-redd}, and ConfBench evaluates confidence for document extraction and review routing \citep{roy-etal-2026-confbench}. Park et al. distinguish error risk from the value of intervention under bounded review budgets \citep{park-etal-2026-risk}. Our question is which error events change residual benchmark error when the repair actions and budget are fixed; under ideal repair, these events are $E_i\subseteq\tau_i(\Pl)$.

Structured calibration estimates probabilities of output events \citep{kuleshov-liang-2015-calibrated,jagannatha-yu-2020-calibrating}. Decision-oriented approaches include property elicitation \citep{frongillo-kash-2021-elicitation}, decision calibration \citep{zhao-etal-2021-calibrating}, efficient multiclass calibration \citep{gopalan-etal-2024-computationally}, and low-bias high-dimensional forecasts with regret guarantees \citep{noarov-etal-2025-high}. Risk control and conformal methods provide predictive guarantees \citep{angelopoulos-etal-2025-learn,angelopoulos-etal-2024-conformal,gui-etal-2024-conformal}, including factuality guarantees for language-model outputs \citep{mohri-hashimoto-2024-language,rubin-toles-etal-2025-conformal}. Prediction-powered inference and betting-based estimation address population quantities \citep{angelopoulos-etal-2023-prediction,waudby-smith-ramdas-2024-estimating}.

\paragraph{Benchmark labels and budgeted evaluation}
Analyses of DocRED have documented missing relations and motivated revised annotations \citep{huang-etal-2022-does,tan-etal-2022-revisiting,heng-etal-2026-re2docred}. Annotation error detection \citep{klie-etal-2023-annotation} and active testing \citep{kossen-etal-2021-active} study how to identify label problems and evaluate systems with limited annotation.

\paragraph{Information acquisition and computation}
Value-of-information methods include estimators of the expected value of sample information \citep{strong-etal-2015-estimating} and knowledge-gradient measurement policies \citep{frazier-etal-2009-knowledge}. Applications span graphical models \citep{krause-guestrin-2005-near,krause-guestrin-2009-optimal}, active decision making and combinatorial exploration \citep{javdani-etal-2014-near,chen-etal-2014-combinatorial}, and information acquisition in extraction or human--agent interaction \citep{narasimhan-etal-2016-improving,dong-etal-2026-value}. Adaptive submodularity addresses sequential selection \citep{golovin-krause-2010-adaptive}.

Model cascades and selective verification allocate computation by expected benefit \citep{chen-etal-2023-frugalgpt,aggarwal-etal-2024-automix,dip-etal-2026-think}. This allocation faces heterogeneous verification signals and uncertain model evaluations \citep{yang-2026-heteroskedastic,khosravi-huo-2026-which}.

\section{Discussion and conclusion}
\label{sec:implications}

Repairability depends on which errors an action can fix together. With true error laws, $\Dcl$ bounds the advantage over scalar-preserving first-order planning. A marginal-risk queue credits node reviews for facts that also require another repair, and can misrank candidates even when $\Dcl=0$. It leaves more errors on both held-out cross-document graphs. Independent-block planning avoids that accounting error, but on DWIE can still lose to a fact-only reference through worse fact ranking. Comparisons should retain fact-only, factorised and joint plans under the same costs, measuring shared-action gains against each model's own fact-only plan.

At a 10\% budget, the fact-scalar queues have precision@budget of 98.8\% and 98.5\%, leaving gaps of 10 and 13 errors to the fact-only oracle. This leaves little audit benefit for better ranking despite AUROC improvements. A generic check improves on verbal confidence at every tested development budget, while its comparison with the label-trained baseline fails the held-out all-comparisons rule. Qwen's larger fact-only gaps at 20\% coexist with smaller verification gains, so the available ranking improvement alone does not predict the value of a check.

Labels determine which predictions enter the audit population and which repairs count as successful. The lenient/strict sensitivity changes the measured value of entity reviews, and the adjudication-based precision calculations depend on accepting LLM judgments. Reports of audit gains should therefore specify the alignment target and unverifiable coverage alongside the actions, budget, harmful decision changes and cost.

\section*{Limitations}
\label{sec:limitations}

\paragraph{Dataset scope}
Confirmation covers two extractors on DocRED-family data. SciREX and DWIE fall below the structural-screen thresholds, and the DWIE policy comparison uses training-sample cross-validation (Section~\ref{sec:development}). Development precision on verifiable predictions is 0.27--0.53, and the higher-precision sweep keeps fitted scores fixed.

\paragraph{Label evidence}
Benchmark-error analysis uses LLM adjudication, with a reference judge from the GPT-4.1 extractor family. Audit populations exclude predictions with unaligned endpoints; cross-document comparisons retain original link labels, and within-document comparisons retain heuristic merge labels. Judge-based sensitivity analyses cover only the reported subsets and oracle quantities.

\paragraph{Verifier and estimation effects}
Each extractor has one default verifier, so extractor and verifier effects are confounded. The development cross-family check is not a full crossed comparison. Bootstrap intervals condition on fitted models and omit training uncertainty; model requests use undated gateway identifiers. Selection-boundary crossing probabilities are not estimated.

\paragraph{Audit assumptions}
The theoretical bounds assume ideal repair, non-adaptive plans, an additive objective and population error laws. The execution-failure sensitivity holds plans fixed and models neither human accuracy nor new reviewer-induced errors. The node-marginal queue is an analytical baseline for marginal-risk accounting, not a reproduction of a published audit tool.

\section*{Ethics statement}
We use DocRED/Re-DocRED (MIT), Linked-DocRED and DWIE (GPLv3), Re$^2$-DocRED (CC BY-SA 4.0) and SciREX (Apache-2.0), under the licences stated in their repositories. Dataset text is excluded from the source release. The main practical risk is treating confidence scores or automated adjudications as sufficient grounds for trusting a graph.

\label{end:body}
\clearpage
\section*{Acknowledgment of AI-generated content}
OpenAI Codex assisted with manuscript revision, LaTeX preparation, figure and table formatting, and consistency checks against experimental results. It also generated the post-hoc repair-failure calculation in Section~\ref{sec:practical}, the driver for the DWIE policy comparison in Section~\ref{sec:development}, and the reproduction scripts. The repair-failure calculation was checked against exact small-instance enumeration, and the reproduction scripts against reference numerical outputs. Figure~\ref{fig:overview} was generated with OpenAI's image-generation tool from an author-specified diagram brief and revised after checking the graph topology; the interface did not expose the backend model version. The action lattice and quantitative plots are rendered from TikZ and numerical data. The study also uses LLMs for extraction, entity linking, verification and exploratory adjudication, with their roles and identifiers specified in Sections~\ref{sec:setup}, \ref{sec:verif} and~\ref{sec:bench}.

\newcounter{preprintbibcount}
\let\preprintbibitem\bibitem
\renewcommand{\bibitem}{\stepcounter{preprintbibcount}%
\ifnum\value{preprintbibcount}=56\relax\newpage\fi\preprintbibitem}
\bibliographystyle{IEEEtranN}
\bibliography{refs,bibliography_settings}
\end{document}